\documentclass[10pt,twocolumn,letterpaper]{article}

\usepackage{wacv}
\usepackage{times}
\usepackage{epsfig}
\usepackage{graphicx}
\usepackage{amsmath}
\usepackage{amssymb}

\usepackage{rotating}
\usepackage{mathtools}
\usepackage{gensymb}
\usepackage{pifont}
\DeclarePairedDelimiter\floor{\lfloor}{\rfloor}

\usepackage{authblk}

\def\wacvPaperID{768} % Enter the WACV Paper ID here

\wacvfinalcopy % *** Uncomment this line for the final submission

\ifwacvfinal
\def\assignedStartPage{1} % *** Enter the assigned starting page number (instead of 9876)
\fi

\ifwacvfinal
\else
\fi

\begin{document}

%%%%%%%%% TITLE
\title{Improving Point Cloud Semantic Segmentation by Learning 3D Object Detection}


\author[1]{Ozan Unal}
\author[1,2]{Luc Van Gool}
\author[1]{Dengxin Dai}
\affil[1]{Computer Vision Lab, ETH Zurich}
\affil[2]{VISICS, ESAT/PSI, KU Leuven}
\affil[ ]{\tt\small \{ozan.unal, vangool, dai\}@vision.ee.ethz.ch}
\date{}
\maketitle
% \thispagestyle{empty}

%%%%%%%%% ABSTRACT
\begin{abstract}
   Point cloud semantic segmentation plays an essential role in autonomous driving, providing vital information about drivable surfaces and nearby objects that can aid higher level tasks such as path planning and collision avoidance. While current 3D semantic segmentation networks focus on convolutional architectures that perform great for well represented classes, they show a significant drop in performance for underrepresented classes that share similar geometric features.  We propose a novel Detection Aware 3D Semantic Segmentation (DASS) framework that explicitly leverages localization features from an auxiliary 3D object detection task. By utilizing multitask training, the shared feature representation of the network is guided to be aware of per class detection features that aid tackling the differentiation of geometrically similar classes. We additionally provide a pipeline that uses DASS to generate high recall proposals for existing 2-stage detectors and demonstrate that the added supervisory signal can be used to improve 3D orientation estimation capabilities. Extensive experiments on both the SemanticKITTI and KITTI object datasets show that DASS can improve 3D semantic segmentation results of geometrically similar classes up to $37.8\%$ IoU in image FOV while maintaining high precision bird's-eye view (BEV) detection results.
\end{abstract}

%%%%%%%%% BODY TEXT

\section{Introduction}
	
The goal of a truly autonomous vehicle is to provide a safer option of travel by removing the human element from the equation. However, this is not trivially accomplished as the vehicle needs to go beyond following a list of rules of the road and complete high level tasks such as path planning and collision avoidance in real time, which not only benefit from knowing the semantics of its immediate surrounding scene including drivable spaces but also the locations of nearby objects. It is therefore crucial for an established 3D semantic segmentation network to correctly identify and segment foreground object classes such as vehicles with high accuracy.

Point cloud semantic segmentation still remains to be a challenging and computationally expensive task. Investigating the current literature we draw the following observations: (1) While 3D semantic segmentation networks perform well for highly represented classes, they show a rapid decrease in performance for underrepresented classes that share similar geometric features. A common example for such a pairing is the car-truck categories, where due to the similarity in their geometric properties, truck segmentation often underperforms because of extensive false negative rates. (2) 3D object detection frameworks perform well for generating high precision 3D bounding boxes while also being capable of differentiation between the common foreground objects. For example in 3D vehicle detection, often the car and truck classes are treated as separate categories in commonly used datasets~\cite{kitti,lyft,nuscenes,argoverse}, thus networks must extract class specific features to correctly classify the objects.

We therefore argue that 3D object detection as an auxiliary task can help improve 3D semantic segmentation results for foreground classes that are underrepresented. Here we will demonstrate, utilizing a car detection auxiliary task can have great benefits when segmenting classes such as trucks or other vehicles. However, in order to utilize both tasks in a unified system in terms of joint supervised training, a dataset is required that contains both supervisory signals. While almost all datasets for 3D object detection lack 3D semantic labels~\cite{kitti, lyft, argoverse, waymo}, those that contain both annotations lack a preestablished benchmark for performance comparison~\cite{audi, pandaset, nuscenes}.

To this end we propose a novel network that we call Detection Aware 3D Semantic Segmentation (DASS), a framework for 3D semantic segmentation that utilizes 3D object detection as an auxiliary task to improve its segmentation performance. Our proposed framework directly consumes irregular point clouds via a PointNet++~\cite{pointnet++} feature extractor to predict semantic labels for points that fall on the front view camera field-of-view (image FOV) while also generating high recall object proposals. DASS is trained using supervisory signals from two partial datasets~\cite{kitti, semantickitti} that only contain a set of annotations for a single task and shows improvements of incredible margins for categories geometrically related to the detection class.

Our key contributions can be summaries as follows: (1) We introduce DASS, a framework for joint point cloud semantic segmentation and 3D object proposal generation from partial datasets. (2) We show that the 3D object detection auxiliary task can improve the generalizability of the shared feature space, enabling vast improvements in 3D semantic segmentation with results up by $37.8\%$ intersection-over-union (IoU) for categories that share geometric features with the detected class. (3) We introduce no additional memory or computational cost compared to a baseline PointNet++ 3D semantic segmentation network~\cite{pointnet++}, as the auxiliary head can be detached during inference. (4) We demonstrate that our proposed network can be used to generate high recall proposals for existing 2-stage 3D object detectors. Overcoming the capacity limitations of multitask training through a novel semantic feature fusion (SFF) connection, our proposed framework shows comparable birds-eye-view (BEV) detection and improved 3D orientation results when used with the second stage of PointRCNN~\cite{pointrcnn}. Furthermore, the resulting network maintains real time inference at a 11Hz rate with only an added $0.15\%$ memory cost, while simultaneously generating accurate 19-class 3D semantic masks.
\section{Related Work}
\label{sec:relatedwork}

In this section, we consider the current approaches for 3D semantic segmentation and 3D object detection. Furthermore we briefly investigate existing multitask learning methods.

\noindent  \textbf{3D Semantic Segmentation:} Point cloud semantic segmentation remains to be a challenging and computationally expensive task in literature. Current benchmarks are dominated mainly by convolutional architectures that project the point cloud onto various representations including spherical representations~\cite{squeezeseg, squeezesegv2}, range images~\cite{rangeNet++}, BEV~\cite{salsanext} and learned 2D representations~\cite{mininet}.

DASS does not utilize a convolutional architecture but is built on a PointNet++ backbone~\cite{pointnet++}. PointNet~\cite{pointnet, pointnet++} proposed a framework that directly consumes point clouds as opposed to parsing to a highly sparse voxel space. \cite{liu2019relation, thomas2019kpconv,liu2019densepoint} extend regular grid CNNs to irregular point configurations by utilizing novel point convolutions to extract local and global features without the loss of information inherent in the process of voxelization. While PointNet based methods tend to underperform~\cite{semantickitti}, this enables DASS to establish encoder level interactions with existing PointNet based 3D object detection frameworks, enabling it to outperform existing convolutional benchmarks.

\noindent \textbf{3D Object Detection:} State-of-the-art 3D object detectors utilize various strategies to deal with the irregular format of point clouds in order to regress the 7 degrees of freedom of a 3D bounding box. Some methods utilize a single stage design, where the final bounding boxes are directly regressed~\cite{second,sassd,3dssd, votenet}, while others prefer a two stage design, where the first stage generates coarse predictions using a region-proposal-network (RPN) and the second stage refines the proposals for the final predictions~\cite{pointrcnn, frustumpointnet, fastpointrcnn, pvpointrcnn}.

DASS utilizes an auxiliary task of 3D object proposal generation. Trained with 3D semantic segmentation, the proposed network provides high recall proposals and thus can be used as a replacement RPN for the currently existing 2-stage architectures like PointRCNN~\cite{pointrcnn} to simultaneously generate 3D detection results with semantic labels.

Compared to the first stage of PointRCNN~\cite{pointrcnn}, DASS completes semantic segmentation of 19 classes while overcoming performance drops in detection that originate from multitask learning. Compared to PointRCNN~\cite{pointrcnn} that predicts binary masks for foreground points which provide explicit information to aid localization, DASS also exploits the unexplored semantic information within the surrounding scene to add additional constraints on the distribution of the bounding boxes, which help further improve orientation estimation.

\noindent  \textbf{Multitask Learning:} Multitask learning (MTL) aims to leverage the supervisory signals of multiple tasks to improve the generalization capabilities of a model. It achieves this by utilizing encoder-level interactions to generate a shared representation~\cite{crossstitch, nddr, mtan}, by using decoder-level interactions to improve single task results from multi-modal distillation~\cite{padnet, papnet}, or a set combination of both.

\cite{dense_mtl} shows that in an MTL setting, performance strongly varies depending on a wide range of parameters (e.g task type, label source) and thus architecture and optimization strategies must be selected on a per case basis. In general it is observed that encoder level interactions perform well for multiple classification problems while decoder level interactions have an advantage in dense prediction tasks.

While MTL has been tackled before in various task types~\cite{maskrcnn, mmf, mtl_depth, meyer2019sensor}, DASS explores the yet unexplored setting of MTL with point cloud semantic segmentation and 3D object detection from two sources of point cloud data that are partially annotated. Through encoder-level interactions, DASS maintains the required inference time and memory for point cloud semantic segmentation. Exploiting decoder level interactions via an SFF layer allows DASS to overcome the performance limitations of encoder-focused MTL and maintain high precision regression for 3D detection.
\begin{figure*}[t]
\centering
\includegraphics[width=\textwidth]{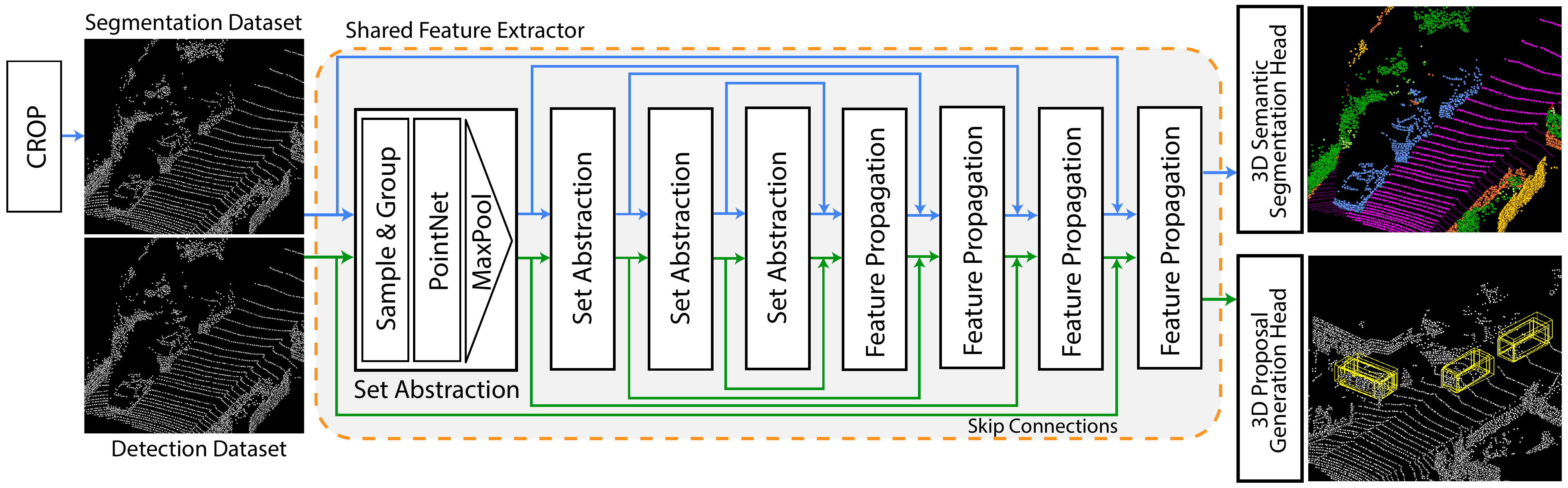}
\caption{Network overview. The network is trained on two partial datasets: (1) with pointwise semantic labels~\cite{semantickitti} and (2) with 3D object annotations~\cite{kitti}. Since the detection datasets provides annotations for just the image FOV, the segmentation dataset is cropped to avoid introducing further domain shifts. A PointNet++ feature extractor is trained on both supervisory signals followed by the individual task heads for the 3D semantic segmentation task and the auxiliary 3D proposal generation task. Best viewed in color.}
\label{fig:network}
\end{figure*}

\section{DASS: Detection Aware 3D Semantic Segmentation}
\label{sec:dass}

In this section we present DASS, a 3D semantic segmentation network with auxiliary 3D object proposal generation, and its training procedure from partial datasets. The overall network pipeline is shown on Fig.~\ref{fig:network}. 

\subsection{Utilizing a Shared Feature Space from Partial Datasets}

We start by defining the tasks of 3D semantic segmentation and 3D object proposal generation.

\noindent \textbf{3D Semantic Segmentation:} Point cloud semantic segmentation is the function $\phi_{seg}$ that assigns a set of semantic labels $L \in \mathbb{Z}^n$, for each point in a given point cloud $P \in \mathbb{R}^{(n \times d)}$ with $n$ points of $d$ dimensions, i.e. ${\phi_{seg}:P \mapsto L}$.

\noindent \textbf{3D Object Proposal Generation:} A bounding box for object $o$ is represented by its 7 degrees of freedom $(x_o,y_o,z_o,h_o,w_o,l_o,r_o)~\in~\mathbb{R}^{7}$, where $(x_o,y_o,z_o)$ define the box center, $(h_o,w_o,l_o)$ define the box height, width, and length respectively, and $r_o$ defines the object rotation around the y-axis in the camera coordinate system. In most autonomous driving cases, the pitch and roll are assumed to be zero. In essence, 3D object proposal generation is the function $\phi_{det}$ that generates $k$ number of bounding boxes within a scene, i.e. returns the set of object bounding boxes $B \in \mathbb{R}^{(k \times 7)}$ for a given point cloud $P$, and proposal count $k$ with $\phi_{det}:P \mapsto B$.

\noindent \textbf{MTL from Partial Datasets:} The goal of multitask training with 3D semantic segmentation and 3D object proposal generation is to find a function $\phi_{MT}$ that returns both per point semantic labels and 3D bounding boxes, i.e. ${\phi_{MT}:P \mapsto (\phi_{seg}(P), \phi_{det}(P))}$.

To learn the mapping of a point cloud $P$ onto a target tuple $(B, L)$ via supervised learning, a dataset is required that contains both ground truth semantic labels $L$ and object bounding boxes $B$. However, amongst the datasets with annotated ground truth bounding boxes, most do not contain per point semantic labels~\cite{kitti, waymo, lyft, argoverse}, while those that do lack an established benchmark for performance comparison~\cite{audi, pandaset, nuscenes}. Thus we are bound to two datasets, each with partial supervisory signals~\cite{kitti, semantickitti}. 

DASS utilizes a shared feature extractor to project the point clouds onto a shared representation $F$. The functions $\phi_{seg}$ and $\phi_{det}$ can then be individually trained with the input and target tuples $(F(P_{seg}),y_{seg})$ and $(F(P_{det}),y_{det})$ respectively, with $P$ the input point cloud and $y$ the target labels, yielding an overall mapping of ${\phi_{MT}~:~P~\mapsto~((\phi_{seg}\circ F)(P),(\phi_{det}\circ F)(P))}$.

In other words, DASS exploits the commonality of 3D semantic segmentation and 3D object detection by utilizing their supervisory signals in parallel. As seen in Fig.~\ref{fig:network}, a PointNet++~\cite{pointnet++} feature extractor is forced to share weights between the primary and auxiliary task, with optimizer steps during training taken from a joint multitask loss. The benefits of such encoder-level interactions when multitask training from partial datasets are three fold: (1) The effective size of the dataset is increased; (2) The training signals of each task act as an inductive bias for the other, improving the generalization capabilities of the model. The feature vector for each point is forced to contain valuable information about its semantic context as well as the detected object class, enhancing segmentation capabilities by allowing better differentiation between geometrically similar classes; (3) By having the bulk of the network parameters reside in the shared feature extractor, the computational overhead and memory requirement of incorporating an additional task is drastically reduced during training. It is also important to note that during inference, the proposal generation head can be detached thus adding no additional cost. DASS can therefore maintain high inference rates while producing accurate semantic masks and 3D object proposals.

\subsection{Joint Proposal Generation and Point Cloud Semantic Segmentation}

As seen in Fig.~\ref{fig:network}, following the shared encoder-decoder, a proposal generation head $\phi_{det}$ and a semanic segmentation head $\phi_{seg}$ are appended to generate 3D semantic labels with coarse detection results. Every batch consists of two mini batches, each with the data from a single partial dataset. After iterative forward passes, a single backward pass is done from the accumulated gradients. In other words, the shared feature extractor is trained from both partial datasets, while the individual heads are trained from a single partial dataset. The shared feature extractor is trained using a multitask loss function given by the weighted sum of the individual task losses of each head, i.e. the first stage total loss is computed as: 
\begin{equation}
\begin{split}
    \label{eq:stage-1}
    \mathcal{L} = & w_{seg} \, \mathcal{L}_{seg}((\phi_{seg} \circ F)(P_{seg}),y_{seg}) \\ 
    & + w_{det} \, \mathcal{L}_{det}((\phi_{det} \circ F)(P_{det}),y_{det})
\end{split}
\end{equation}
with $w_{seg}$, $\mathcal{L}_{seg}$ denoting the 3D semantic segmentation weight and loss and $w_{det}$, $\mathcal{L}_{det}$ denoting the 3D object detection weight and loss.

\noindent \textbf{Segmentation Loss:} The 3D semantic segmentation head seen in Fig.~\ref{fig:network} generates pointwise semantic labels. The head is trained using the cross-entropy loss with a weight vector $w_{classes}$ to deal with the class imbalance. The segmentation loss is given by
\begin{equation}
    \label{eq:seg_loss}
    \mathcal{L}_{seg} = \mathcal{L}_{CE}(L, \hat L ; w_{classes})
\end{equation}
with  $\mathcal{L}_{CE}$ denoting the cross entropy loss, $L$ and $\hat L$ denoting the sets of estimated semantic labels and their corresponding ground truths respectively.

% \begin{figure}[t]
%     \centering
%     \includegraphics[width=0.40\textwidth]{figures/reg_loss.pdf}
%     \caption{Illustration of the per-point anchor free bin-based localization.}
%     \label{fig:reg_loss}
% \end{figure}

\noindent \textbf{Detection Loss:} For the auxiliary 3D proposal generation task, we use a per-point bin based loss function following PointRCNN~\cite{pointrcnn}. The surrounding area of each point is discretized into a set number of bins. This allows us to restate the problem of the bounding box center localization on the transverse plane $(x,z)$ as a classification problem which are shown to be better fitted for encoder-focused architectures~\cite{dense_mtl}. To achieve finer details, we allow a residual to be regressed for each bin. For a bin size of $\delta$ in a surrounding area of radius $S$, we define $k = \delta \, (k_{bin} + 1/2) + k_{res} - S$ for $k \in \{x,z\}$. Here $k_{bin} \in [\floor{-S/\delta}, ..., -1, 0, 1, ..., \floor{S/\delta}]$ defines the bin in which the target is located, and $k_{res} \in (-\delta/2, \delta/2)$ defines the residual within that bin. Similarly, the rotation estimation is also restated as a classification problem where the transverse plane is divided into a set number of angles $\alpha$ where again we define $k = \alpha \, k_{bin} + k_{res}$ for $k \in \{r\}$ with $k_{bin} \in [0, 1, ..., \floor{2 \pi / \alpha}]$ and $k_{res} \in (-\alpha/2, \alpha/2)$.

As all objects within a driving scene are ground bound and have similar sizes per class, the elevations  $y$ and size residuals $(h,w,l)$ of bounding boxes follow very narrow distributions, allowing these values to be regressed directly. The resulting loss function is given by:
\begin{equation} 
\begin{split}
    \label{eq:det_loss}
    \mathcal{L}_{det} = & \sum_{k \in \{x,z,r\}} \left ( \mathcal{L}_{CE}(k_{bin}, \hat{k}_{bin}) + \mathcal{L}_{sL1}(k_{res}, \hat{k}_{res}) \right ) \\ & + \sum_{j \in \{y,h,w,l\}} \mathcal{L}_{sL1}(j,\hat j)
\end{split}
\end{equation}
where hat denotes the ground truth and $\mathcal{L}_{sL1}$ denotes the smooth L1 loss.

All points that lie outside of ground truth bounding boxes do not contribute to the detection loss.

\subsection{DASS in a 2-stage 3D Object Detection Pipeline}
\label{sec:2stage}

\begin{figure}[t]
\centering
\includegraphics[width=0.45\textwidth]{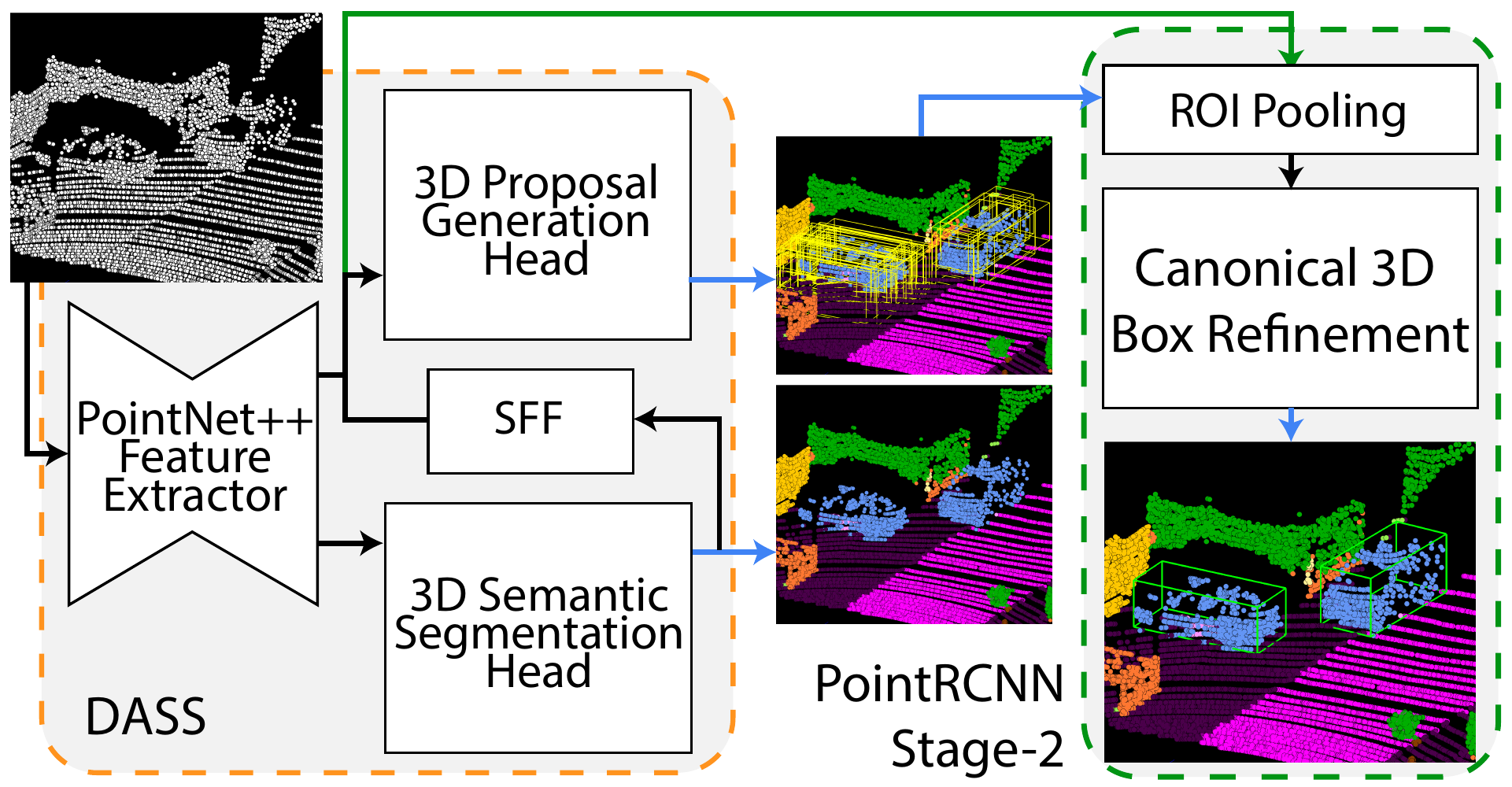}
\caption{Network extension overview. DASS is used as the RPN of PointRCNN~\cite{pointrcnn}. To further improve proposal generation results semantic feature fusion (SFF) is applied before the proposal generation.}
\label{fig:extension}
\end{figure}

We make the following observation: Each semantic mask can act as a constraint on a bounding boxes distribution. For example cars are likely to leave substantial space between themselves and surrounding buildings; cars, cyclists and pedestrians are more likely to be rotated along the direction of the road or sidewalks respectively and away from paths of direct collision~\cite{neverwalkalone}. We therefore argue that, not only does the  3D semantic segmentation task benefit from the auxiliary 3D proposal generation, but the opposite also holds especially for BEV detection and 3D orientation. The 3D proposal generator can therefore benefit highly from knowing the semantic masks of specific background classes, thus DASS can be used to generate high recall proposals for existing 2-stage detectors.

In Fig.~\ref{fig:extension}, we illustrate a pipeline for PointRCNN~\cite{pointrcnn} that utilizes DASS as the primary proposal generator. The proposals are initially expanded to form regions of interests which are then pooled with the shared features to generate the input of the second stage. The second stage applies canonical box refinement on the generated proposals to predict the final 3D detection results~\cite{pointrcnn}.

With the proposed extension, DASS becomes the first pipeline to generate 19-class point cloud semantic segmentation results with 3D object bounding boxes in real time at an operating frequency of 11Hz (Nvidia Titan Xp 12G GPU 2.2GHz) with a minimal added memory cost of just $0.15\%$. This can be highly beneficial when dealing with real world problems of navigation in complex scenes that involves on-the-go path planning and collision avoidance.

\noindent \textbf{Semantic Feature Fusion:} In MTL, the problem of invariance vs. sensitivity is always apparent~\cite{dense_mtl}. The network may never converge to a state where it extracts a vital feature for one task because it contradicts with the objective of the other. With DASS, we tackle a currently unexplored area of MTL and consider two tasks that directly consume point clouds with 3D object detection and 3D semantic segmentation from partial datasets. While 3D semantic segmentation performs well under the limited capacity of the feature extractor, we observe a severe reduction in performance for high precision localization.

To overcome this issue, we introduce semantic feature fusion (SFF) as a decoder level interaction as seen in Fig.~\ref{fig:network}. During training, with gradient accumulation set to zero, we do a forward pass of the detection mini-batch through the shared feature extractor and the segmentation head to infer the per point semantic label likelihoods. SFF learns to summarize this high dimensional likelihood vector through 1D convolutional operations and provides the 3D proposal generation head a compact representation of the scene semantics. These are concatenated back with the shared features to be input to the 3D proposal generation head. By directly utilizing such compacted information, we minimize adding redundant information and further complexity to the system, while still allowing the detection head to directly extract inter class dependencies. With SFF, DASS achieves higher 3D detection recall across all thresholds, which is the goal of the first stage detector as precise localization is carried out in the second stage.
\section{Experiments}
\label{sec:experiments}

In this section we look into the architecture specifications and training scheme of DASS from partial datasets. We report the results of our network and provide ablation studies on the training procedure and individual components.

\subsection{Network Architecture}

For the PointNet++~\cite{pointnet++} encoder-decoder, 4 set-abstraction layers with multi-scale grouping are used with group sizes of 4096, 1024, 256, 64 points of increasing radii. Every group \& sampling operations of the set abstraction layers are followed by a block of 3 linear layers for each of the two scales. The set abstraction layers feed into 4 feature propagation layers with skip connections to obtain per point feature vectors that are rich in both semantic and class-specific information. While using 2-scale grouping allows us to introduce scale invariance into our network, the hierarchical structure of the PointNet++ feature extractor captures better local properties which benefit both tasks.

Both the first stage 3D semantic segmentation and the 3D object proposal generation heads consist of a single 1D convolutional layer of size 128. Batch norm and ReLU activation is applied after every layer. The learning rate is set to 0.002. Adam optimizer is used with a one-cycle learning rate scheme. The weight decay is set to 0.001 with momentum of 0.9.

\subsection{Training Scheme}

\noindent \textbf{Dataset:} Due to a lack of unified dataset that contains both ground truths, two partial datasets are utilized for training a weight-sharing PointNet++ encoder-decoder structure~\cite{pointnet++}, such that the resulting pointwise feature vectors include both object-class and semantic information. In specific, the 3D semantic segmentation pipeline is trained on the SemanticKITTI \textit{train} set~\cite{semantickitti} while the 3D object proposal network is trained on the KITTI \textit{train} set~\cite{kitti} with the car category.

\noindent \textbf{Point Cloud Preprocessing and Augmentation: } As KITTI~\cite{kitti} does not provide ground truth 3D bounding box annotations for the full $360 \degree$ view, we crop the 3D point clouds such that all points lie within the image FOV. Applying the same transformations to both partial datasets is crucial as this avoids added domain shifts within the training process and thus prevents overfitting to domain specific features. This in turn means that our semantic segmentation network operates exclusively on image FOV despite the provided $360\degree$ labels from SemanticKITTI~\cite{semantickitti}. Each cropped region is then randomly subsampled to contain a 16384 points. If a scene contains less points after the cropping it is zero-padded until the fixed value is reached. The reflectance intensity values of all points within the point cloud are normalized by subtracting 0.5. 

It is important to note that given a dataset that contains $360 \degree$ annotations for both tasks, our method can easily be reapplied to provide $360 \degree$ semantic labels and will again show the same benefits of adding no memory or computational cost during inference.

Two data augmentation schemes are implemented. (1) For both datasets, each scene is randomly rotated by an angle sampled from $[-10\degree, 10\degree]$, scaled by a scale factor sampled from $[0.95, 1.05]$ and flipped randomly with 0.5 probability. (2) For the object detection dataset only, following~\cite{second,pointrcnn} ground truth bounding box augmentation is applied, where ground truth boxes and their respective points are taken from various scenes to be implanted in another. The new box and its points are placed within the point cloud at the same location assuming that there is no existing overlapping box. The box with its points are then translated such that it lies on the ground plane. Furthermore, points above and below the box are removed from the point cloud. %Ground truth data augmentation~\cite{pointrcnn} is applied to all scenes with a hard-ratio of 0.6, meaning a sample is selected from [40, 80] meters more often than an easy one to improve the performance of the network for distant vehicles.

\noindent \textbf{Training:} A mini-batch size of 8 is used for both tasks which yields cleaner gradients compared to an unbalanced grouping strategy. Due to the size differences of the two datasets, we define an epoch for the first stage as a single iteration over the SemanticKITTI~\cite{semantickitti} dataset with multiple shuffled cycles over the KITTI object dataset~\cite{kitti}. The two tasks are weighed by (1.5, 1) for 3D semantic segmentation and 3D object detection respectively following the inverse of their converged individual losses. The network is trained for 75 epochs.

For the training of the stage-1 regression head, only the points that lie inside a bounding box are considered in the loss function. 

Following~\cite{pointpillars}, all car objects that lie outside of the range x, y, z [[-40, 40], [-1, 3], [0, 70.4]] meters are filtered out. The mean car bounding box is set to have a height, width and length of (1.5, 1.6, 3.9) meters and the size of each object $(h,w,l)$ is regressed as the difference from the mean anchor. The search scope is set to 3.0$m$ resulting in a bin size of $\delta = 0.5m$ with 6 equal length bins, and the rotation range is set to $2\pi$ with 12 discrete heads. Vans are also considered within the car category.

To deal with the class imbalance apparent in the SemanticKITTI~\cite{semantickitti} dataset, weighted cross entropy is used with a 19 class weight vector $w_{classes}$ given by the inverse of a class' frequency within the entire set of point clouds.

\begin{table}
\vspace{-3mm}
\begin{center}
\tabcolsep=0.11cm
\rule{0pt}{4ex} \\
\resizebox{0.49\textwidth}{!}{\begin{tabular}{|l| c c c c|} 
\hline
 & Parameters & Inference & 3D Semantic & 3D Object\\
Method & [Million] & Time [s]&  Segmentation & Detection\\
 \hline
SqueezeSeg\cite{squeezeseg} &1 &0.015& 360\degree & - \\ 
SqueezeSegV2\cite{squeezesegv2} & 1& 0.02& 360\degree & -\\ 
DarkNet21\cite{rangeNet++} & 25 &  0.055& 360\degree & -\\ 
DarkNet52\cite{rangeNet++} & 50 &  0.01& 360\degree  & -\\ 
PointRCNN & 3.90 & 0.09 & - & \ding{51} \\ 
\hline
Semantic Baseline & 3.04 &  0.015& Image FOV  & - \\
DASS & 3.04 &  0.015& Image FOV  & Proposal\\
DASS+RCNN & 3.91 & 0.09 & Image FOV & \ding{51} \\
 \hline
\end{tabular}
}
\caption{\label{tab:stats}Approach summaries.}
\end{center}
\vspace{-6mm}
\end{table}

\subsection{Results}
\begin{figure*}[t]
\centering
\includegraphics[width=\textwidth]{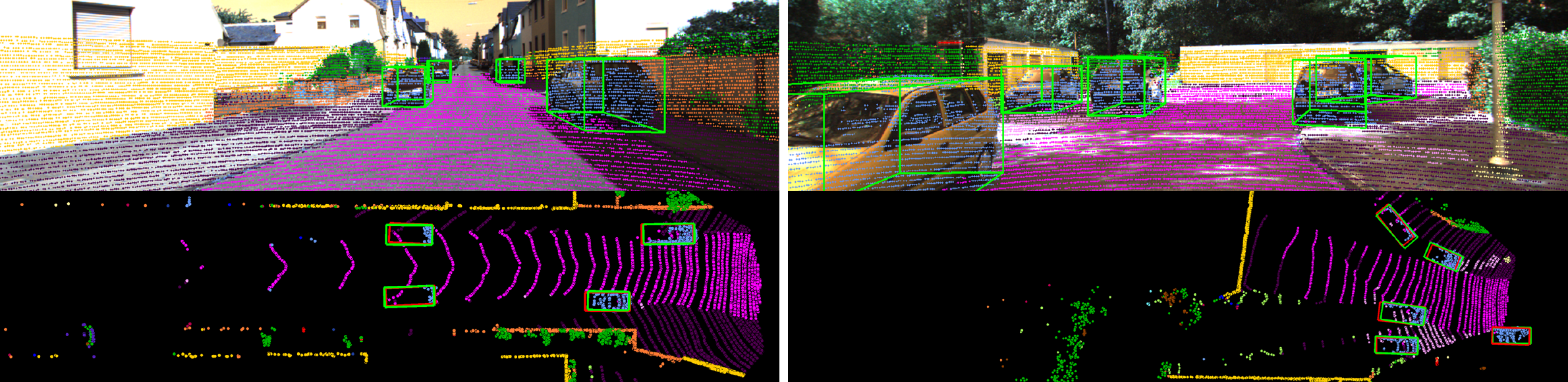}
\caption{Example results from the KITTI~\cite{kitti} \textit{val} set. Shown are (top) pointwise semantic labels and predicted bounding boxes in green overlayed onto the camera 2 image; (bottom) BEV results with ground truth boxes in red and predicted boxes in green. The multitask results are generated using DASS as the RPN for PointRCNN~\cite{pointrcnn}. Best viewed in color.}
\vspace{-8mm}
\label{fig:results}
\end{figure*}

We report 3D semantic segmentation results from the SemanticKITTI~\cite{semantickitti} $val$ set on image FOV and the recall for 3D object detection at varying thresholds from the KITTI~\cite{kitti} $val$ set. We demonstrate on the KITTI~\cite{kitti} $test$ set that our network can be used in conjunction with existing 2-stage detectors to generate comparable BEV detection results, improved 3D orientation estimation, to maintain minimal memory cost, to operate in real time at 11Hz frequency, all while simultaneously generating 3D semantic masks. Further statistics on evaluated approaches can be found on Tab.~\ref{tab:stats}. Example results can be seen in Fig.~\ref{fig:results}.

\begin{figure*}[t]
\centering
\includegraphics[width=\textwidth]{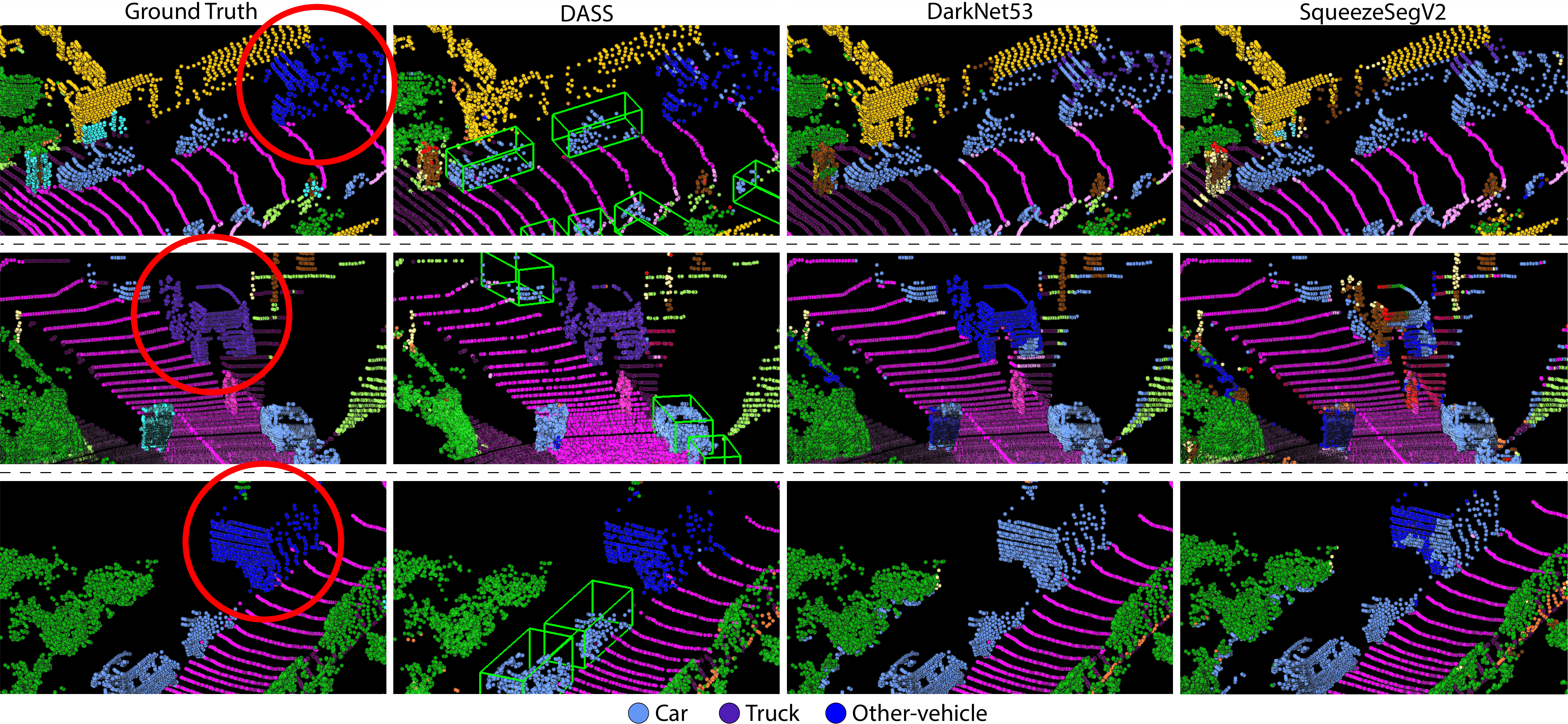}
\caption{Point cloud semantic segmentation results on the SemanticKITTI~\cite{semantickitti} \textit{val} set. DASS is compared against the ground truth labels~\cite{semantickitti}, DarkNet53~\cite{rangeNet++} and SqueezeSegV2~\cite{squeezesegv2}. To illustrate the detection awareness of DASS, we overlay the predicted bounding boxes from the DASS+RCNN network. Here we draw an emphasis on the differentiation capabilities of DASS for the vehicle classes, with corresponding colors and class tags given below the figure. Best viewed in color.}
\vspace{-1mm}
\label{fig:additional_results2}
\end{figure*}
\begin{table*}[t]
   \tabcolsep=0.11cm
\resizebox{\textwidth}{!}{\begin{tabular}{|r|c|ccccccccccccccccccc|}
\hline
\multicolumn{1}{| r |}{Method} & mIoU &\begin{turn}{90} car \end{turn}&\begin{turn}{90} bicycle \end{turn}&  \begin{turn}{90}motorcycle \end{turn}&\begin{turn}{90} truck \end{turn} &\begin{turn}{90} other vehicle \ \ \end{turn} &\begin{turn}{90} person\end{turn} &\begin{turn}{90} bicyclist\end{turn} &\begin{turn}{90} motorcyclist \end{turn} &\begin{turn}{90} road\end{turn} & \begin{turn}{90}parking\end{turn}  &\begin{turn}{90} sidewalk\end{turn} &\begin{turn}{90}other ground\end{turn} &\begin{turn}{90} building \end{turn} &\begin{turn}{90} fence \end{turn}&\begin{turn}{90} vegetation\end{turn} &\begin{turn}{90} trunk\end{turn} &\begin{turn}{90} terrain\end{turn} & \begin{turn}{90}pole\end{turn} & \begin{turn}{90}traffic sign \end{turn}\\[0.5ex] 
\hline

SqueezeSeg~\cite{squeezeseg} & 32.1 & 75.9& 12.8 &11.5& 3.3 &4.0 &22.0& 34.4 &0.1 &91.7 &16.2 &62.7 &0.4 &57.2 &18.8 &67.1& 27.7 &65.6 &23.3 &14.5 \\ 

SqueezeSeg-CRF~\cite{squeezeseg} & 33.4 & 75.1& 13.6 &15.4& 3.9 &10.6 &27.7& 39.9 &\textbf{0.2} &90.8 &16.4 &62.2 &\textbf{0.6} &61.8 &22.7 &66.5& 26.7 &65.9 &13.7 &21.1 \\ 

SqueezeSegV2~\cite{squeezesegv2} & 41.3 & 84.1& 15.1 &24.8& 25.1 &27.9 &23.4& 44.7 &0.0 &94.4 &36.9 &74.3 &0.1 &71.3 &37.7 &74.5& 36.2 &69.8 &21.8 &22.0 \\ 

SqueezeSegV2-CRF~\cite{squeezesegv2} & 42.6 & 85.5& 18.5 &39.3& 23.2 &31.4 &33.7& 54.6 &0.0 &94.3 &32.3 &73.4 &0.2 &69.3 &39.4 &71.6& 37.5 &69.2 &14.8 &20.9 \\ 
 
DarkNet21~\cite{rangeNet++} & 49.0 & 86.5 & 27.6 &39.6& 35.5 &23.4 &43.2& 50.1 &0.0 &\textbf{95.9} &\textbf{40.9} &79.8 &0.0 &77.3 &50.7 &81.4& 53.7 &72.0 &42.2 &31.9 \\ 

Darknet52-512~\cite{rangeNet++} & 34.9 & 79.8& 14.9 &17.0& 3.9 &14.6 &11.2& 26.9 &0.0 &93.3 &21.2 &70.9 &0.1 &59.2 &30.9 &70.6& 36.6 &65.7 &23.8 &23.4 \\ 

DarkNet53-1024~\cite{rangeNet++} & 39.2 & 83.9& 17.5 &3.5& 24.1 &8.3 &10.9& 43.3 &0.0 &94.2 &15.3 &74.1 &0.0 &70.6 &47.7 &78.0& 38.5 &70.0 &34.8 &30.0 \\ 

DarkNet53~\cite{rangeNet++} & 51.0 & 86.9& 26.6 &\textbf{47.5}& 34.0 &27.2 &\textbf{51.6}& 62.7 &0.0 &\textbf{95.9} &39.7 &\textbf{80.0} &0.0 &77.8 &\textbf{51.1} &81.3& 53.6 &71.9 &\textbf{46.7} &33.8 \\ 

\hline

Semantic Baseline & 48.0 & 89.8 & \textbf{33.7} & 29.9 & 28.9 & 28.9 & 35.5 & 62.8 & 0.0 & 94.6 & 32.1 & 75.7 & 0.4 & \textbf{82.1} & 42.6 & 83.3 & 53.0 & \textbf{73.6} & 38.2 & 26.3 \\

DASS & \textbf{51.8} & \textbf{91.4} & 25.8 & 31.0 & \textbf{66.7} & \textbf{43.8} & 47.7 & \textbf{70.8} & 0.0 & 92.8 & 31.7 & 71.0 & 0.0 & \textbf{82.1} & 39.1 & \textbf{83.5} & \textbf{56.6} & 69.6 & 45.5 & \textbf{35.1} \\

\hline 
\end{tabular}
}
\caption{ \label{tab:semantickitti_evaluation} 3D semantic segmentation results. All networks are evaluated on image FOV using the SemanticKITTI~\cite{semantickitti} validation set. Semantic Baseline denotes DASS trained without the auxiliary task.}

\vspace{-3mm}
\end{table*}

\begin{table*}[t]
\begin{minipage}{.38\textwidth}
    \begin{center}
\tabcolsep=0.11cm
\rule{0pt}{4ex} \\
\resizebox{\textwidth}{!}{\begin{tabular}{|r| c c c c c|} 
\hline
 & \multicolumn{5}{c |}{Recall at IoU [\%]} \\
\multicolumn{1}{| r |}{Method} & 0.1 & 0.3 & 0.5 & 0.7 & 0.9 \\  
 \hline
  PointRCNN\cite{pointrcnn} & 96.84 & 95.71 & 93.49 & \textbf{73.37} & \textbf{1.10} \\
  DASS & 96.90 & 96.19 & 93.80 & 68.95 & 0.40 \\
  DASS+SFF & \textbf{97.36} & \textbf{96.45} & \textbf{94.24} & 71.47 & 0.65\\
 \hline
\end{tabular}
}
\caption{\label{tab:recall}Recall results for 3D detection at varying IoU thresholds for the car class. All networks are evaluated on the KITTI~\cite{kitti} \textit{val} set.}
\end{center}
\end{minipage}
\hfill
\begin{minipage}{.57\textwidth}
    
\begin{center}
\tabcolsep=0.11cm
\rule{0pt}{4ex} \\
\resizebox{\textwidth}{!}{\begin{tabular}{|r| c c c|c c c|} 
\hline
 & \multicolumn{3}{c|}{BEV [\%]}& \multicolumn{3}{c|}{Orientation [\%]} \\
Method &  Easy& Moderate & Hard &  Easy& Moderate & Hard\\
 \hline
PointRCNN\cite{pointrcnn} &  \textbf{92.13} & \textbf{87.39} & \textbf{82.72} & 95.90 &	91.77 &	86.92 \\
DASS+RCNN & 91.74 &	85.85 &	80.97 & \textbf{96.20} &	\textbf{92.25} & \textbf{87.26} \\
 \hline
\end{tabular}
}
\caption{\label{tab:od}Evaluation of BEV and 3D orientation on the KITTI~\cite{kitti} \textit{test} set for the car class. As seen DASS can be used as a RPN for PointRCNN~\cite{pointrcnn} (DASS+RCNN) to achieve comparable BEV and 3D orientation results while generating 3D semantic masks ranging 19 classes.}
\end{center}\vspace{-15px}

\end{minipage}
\vspace{-2mm}
\end{table*}
\subsubsection{3D Semantic Segmentation Results}
\label{sec:segmentation_results}
As DASS operates on image FOV, the performance cannot be evaluated on the SemanticKITTI~\cite{semantickitti} \textit{test} set which requires full $360 \degree$ annotations. Thus we opt to use the SemanticKITTI~\cite{semantickitti} \textit{val} set on image FOV, where all results are evaluated using the official metric of per class mean intersection-over-union (mIoU).

The results can be seen on Tab.~\ref{tab:semantickitti_evaluation} where DASS is compared against the benchmark networks provided by SemanticKITTI~\cite{semantickitti}. The per point label predictions of various SemanticKITTI benchmark networks including SqueezeSeg~\cite{squeezeseg}, SqueezeSegV2~\cite{squeezesegv2} and DarkNet~\cite{rangeNet++, semantickitti} have been released. Here, we reevaluate these published results on image FOV. Our proposed network shows an overall better classification performance than the benchmarks with a mIoU improvement of $0.8\%$, with emphasis drawn on classes that share geometric features with the car category. As observed, in the car, truck, and other vehicle categories, DASS outperforms all benchmark networks with incredible margins of $4.5\%, 32.7\%, 16.6\%$ respectively. Due to the added supervisory signals of the 3D object detection task, the task specific head can utilize the shared features to better distinguish classes of similar characteristics.

In Fig.~\ref{fig:additional_results2} we demonstrate this differentiation capability of DASS amongst geometrically similar vehicle classes and compare it to existing benchmark segmentation networks \cite{rangeNet++, squeezesegv2}. We additionally overlay the predicted bounding boxes of DASS+RCNN to illustrate the detection awareness of DASS. Here it is shown that DASS can better separate the truck (purple) and other vehicle (blue) classes from the overrepresented car (light blue) class thanks to its auxiliary detection task.

\subsubsection{3D Object Detection Results}

Here we provide results for the commonly reported \textit{car} category. Following PointRCNN~\cite{pointrcnn}, we initially generate 9000 proposals which are then reduced to 100 using distance based sampling and non-maximum suppression (NMS) with a threshold of 0.8.

The first stage recall values are seen in Tab.~\ref{tab:recall}. Here we observe both the benefits and detriments of multitask training. The increased generalization capabilities of the shared feature extractor allows the object proposal to generate higher recall proposals for lower thresholds of $0.1, 0.3$ and $0.5$ with recall values seeing an increase of $0.52\%, 0.74\%$, and $0.75\%$ respectively. However, the localization performance suffers at higher thresholds as the two tasks compete for capacity. A feature that is highly beneficial for localization may be a nuisance for the 3D semantic segmentation task. Thus at higher thresholds of $0.7, 0.9$, DASS underperforms by $1.9\%,0.45\%$ compared to the stage-1 of PointRCNN.

We use DASS as the stage-1 object proposal generator and append the stage-2 refinement network of PointRCNN to generate BEV and 3D orientation results, as we expect DASS to provide additional auxiliary information that can benefit such tasks (see Sec.~\ref{sec:2stage}). We call this network DASS+RCNN. In Tab.~\ref{tab:od} the KITTI~\cite{kitti} \textit{test} set results are given, however it should be noted that while DASS+RCNN is trained on the official train/val split (50/50), the reported PointRCNN results are obtained with a (80/20) split. Nonetheless our 3D semantic segmentation network still performs at a comparable level with PointRCNN~\cite{pointrcnn}. In the BEV category DASS+RCNN falls just shy on every difficulty by $0.39\%, 1.54\%, 1.75\%$ but achieves better results in 3D orientation by $0.3\%, 0.48\%, 0.34\%$ while simultaneously generating 3D semantic masks ranging 19 classes and only causing an additional $0.15\%$ memory requirement.

\subsection{Ablation Studies}
\label{sec:ablation}
In this section we provide extensive ablation studies to analyze both the effectiveness of the added components. All components are evaluated on the KITTI~\cite{kitti} and SemanticKITTI \textit{val} splits~\cite{semantickitti} for 3D object detection and 3D semantic segmentation respectively.

\noindent \textbf{Auxiliary 3D Object Proposal Generation: } In Tab.~\ref{tab:semantickitti_evaluation} we provide a comparison of our proposed network trained without the aid of the auxiliary task, which we call \textit{Semantic Baseline}. Similar to the comparison drawn in Sec.~\ref{sec:segmentation_results}, we observe an overall increase in mIoU by $3.8\%$, with the performance boost mainly coming from the car, truck and other vehicle classes with mIoU increases of $1.6\%, 37.8\%, 14.9\%$ respectively.

However it should also be noted that some classes show inferior results when multitask training. As stated before, the shared feature space may never converge to a state where a crucial information for 3D semantic segmentation exists, if that feature contradicts with the objectives of the auxiliary detection task. An example can be observed by the drop in performance for the road and parking classes by up to $1.8\%$ and $0.4\%$ respectively. While drivable surfaces provide information regarding the boundaries of the vehicles (e.g. the elevation of the bounding box as it must lie on a drivable surface), the distinction between drivable surfaces (e.g. road vs. parking) does not provide any further information in that regard which results in misidentified region boundaries.

\noindent \textbf{Semantic Feature Fusion}: We evaluate the effectiveness of SFF by drawing comparisons to the baseline DASS. With its increased generalization capabilities, DASS matches or exceeds PointRCNN~\cite{pointrcnn} RPN's 3D recall at lower IoU thresholds. However, as the two tasks compete for capacity, the network fails to extract the much needed task-specific features that aid high precision localization.

As seen in Tab.~\ref{tab:recall}, providing the 3D object proposal head with summarized semantic context via an SFF layer improves its recall for all thresholds. In specific, the detection head mostly benefits from the semantically rich feature at higher IoU thresholds of $0.7$ and $0.9$ with recall increases of $2.52\%$ and $0.25\%$ respectively. This enables the network to achieve better localization by extracting further class-specific and inter-class dependencies.
\section{Conclusion}
\label{sec:conclusion}

In this work we proposed a Detection Aware 3D Semantic Segmentation (DASS) network to tackle limitations of current architectures. Our proposed network utilizes an auxiliary 3D object detection task to guide the shared feature representation into extracting localization features that allow better differentiation between geometrically similar foreground classes. Experiments on the SemanticKITTI dataset show that this significantly improves 3D semantic segmentation in image FOV without any additional memory requirement or computational overhead, as the auxiliary task head can be detached during inference. We further investigate the yet unexplored problem of multitask learning of 3D semantic segmentation and 3D object detection from point clouds. We showcase a 2-stage 3D object detection pipeline that utilizes DASS as a RPN with preexisting architectures and overcome the capacity limitations of multitask learning through semantic feature fusion. Experiments on the KITTI dataset show that DASS can improve 3D orientation estimation while preserving BEV detection results, operating in real time, costing negligible memory, and producing highly accurate 3D semantic masks.

\noindent \textbf{Acknowledgements:} This work was funded by Toyota Motor Europe via the research project TRACE Zurich.

{\small
\bibliographystyle{ieee_fullname}
\bibliography{egbib}
}

\end{document}